\documentclass[10pt,twocolumn,letterpaper]{article}

\usepackage[pagenumbers]{cvpr} % To force page numbers, e.g. for an arXiv version

\usepackage{cite}
\usepackage{graphicx}
\usepackage{amsmath}
\usepackage{amssymb}
\usepackage{booktabs}
\usepackage{svg}
\usepackage{float}
\usepackage{pifont}
\usepackage{diagbox}

\usepackage[pagebackref,breaklinks,colorlinks]{hyperref}

\usepackage[capitalize]{cleveref}
\crefname{section}{Sec.}{Secs.}
\Crefname{section}{Section}{Sections}
\Crefname{table}{Table}{Tables}
\crefname{table}{Tab.}{Tabs.}

\def\cvprPaperID{*****} % *** Enter the CVPR Paper ID here
\def\confName{CVPR}
\def\confYear{2022}

\begin{document}

\title{Mitigating Motion Blur for Robust 3D Baseball \\ Player Pose Modeling for Pitch Analysis}

% \author{Jerrin Bright, Yuhao Chen, John Zelek\\
% Vision and Image Processing Lab, University of Waterloo, Ontario, Canada\\
% { {\{jbright, yuhao.chen1, jzelek\}@uwaterloo.ca}}

\author{Jerrin Bright\\
University of Waterloo\\
Waterloo, Ontario, Canada\\
{\tt\small jerrin.bright@uwaterloo.ca}
% 12/08/2023\\
% { \fontsize{12}{18}{\textbf{SYDE770 - Project Report}\\}}
\and
Yuhao Chen\\
University of Waterloo\\
Waterloo, Ontario, Canada\\
{\tt\small yuhao.chen1@uwaterloo.ca}
\and
John Zelek\\
University of Waterloo\\
Waterloo, Ontario, Canada\\
{\tt\small jzelek@uwaterloo.ca}
}
\maketitle

%%%%%%%%% ABSTRACT
\begin{abstract}
Using videos to analyze pitchers in baseball can play a vital role in strategizing and injury prevention. Computer vision-based pose analysis offers a time-efficient and cost-effective approach. However, the use of accessible broadcast videos, with a 30fps framerate, often results in partial body motion blur during fast actions, limiting the performance of existing pose keypoint estimation models. Previous works have primarily relied on fixed backgrounds, assuming minimal motion differences between frames, or utilized multiview data to address this problem. To this end, we propose a synthetic data augmentation pipeline to enhance the model's capability to deal with the pitcher's blurry actions. In addition, we leverage in-the-wild videos to make our model robust under different real-world conditions and camera positions. By carefully optimizing the augmentation parameters, we observed a notable reduction in the loss by 54.2\% and 36.2\% on the test dataset for 2D and 3D pose estimation respectively. By applying our approach to existing state-of-the-art pose estimators, we demonstrate an average improvement of 29.2\%. The findings highlight the effectiveness of our method in mitigating the challenges posed by motion blur, thereby enhancing the overall quality of pose estimation.
\end{abstract}

\begin{figure}
{\centering
\includegraphics[width=8cm]{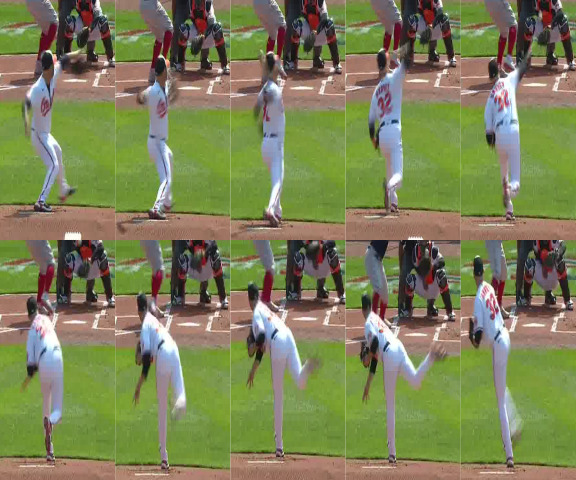}
\caption{Sequence \cite{youtube} captured at 30 frames per second from behind the homeplate view of the pitcher.} 
% while pitching the ball. Notably frames 7 and 8 show a discernible motion blur effect within the pitching hand. The pitching hand is completely occluded as shown in frame 9, adding an extra layer of visual complexity to the captured data. }
\label{fig:dataset}
}
\end{figure}

\section{Introduction}

Sabermetrics, the empirical analytics approach to in-game baseball analysis, has seen remarkable growth in recent times \cite{baseball-review}. While most Sabermetrics work focuses on structured statistical data \cite{baseball1, baseball2} such as pitch type, break, spin rate, and historical win rate; sports video analytics offers the potential for visual understanding, detailed performance assessment, and contextual information of real-time data. Baseball, often regarded as a sport with extensive statistical analysis, provides valuable insights into various aspects of the game and player skills. Being a pitcher-friendly game, the performance of the pitcher significantly influences the team's success and overall gameplay. Analyzing pitchers in baseball is of utmost importance as it can significantly enhance the assessment of pitching techniques, accurately evaluate pitch movements, and aid in detecting subtle patterns, such as changes in delivery or pitch tipping. This comprehensive analysis not only provides valuable information about individual pitchers but significantly contributes to improving the overall performance of the team.

A key challenge in pitcher pose estimation from broadcast videos is the quality of the input image, as factors such as motion blur and self-occlusion can degrade the performance of the reconstruction. In Figure \ref{fig:dataset}, we illustrate an instance of the challenges posed by a substantial motion blur effect during pitching action, coupled with self-occlusion from the homeplate view. These issues underscore the complexity of the task and emphasize the need for robust and sophisticated methods to address such inherent limitations in the data. While some prior works have addressed motion blur caused by the camera \cite{blurcam1, blurcam2, blurcam3}, the problem of human-articulated motion blur remains largely unexplored. The impact of such motion blur on human pose estimation is significant. Significant advances have been made in tackling this issue \cite{zhao2023human, Lumentut_2020_ACCV, holistic-deblur}, however, challenges still persist when dealing with dynamic backgrounds or fast-moving human motions. 

% \cite{frankmocap}

Thus, this paper presents a unique approach for accurate pose estimation of pitchers in baseball games, considering the challenges addressed previously. Unlike existing methods that rely on complex pipelines, we propose a strategy centered on smart augmentation effects. By augmenting the training data with selective motion blur effects, we enhance the network's ability to learn and adapt to these effects. The inclusion of in-the-wild data from the internet significantly bolstered the network's generalization capabilities to different camera positions and lighting conditions. This research highlights the effectiveness of a focused augmentation strategy and challenges the conventional notion of complex pipelines for handling motion blur. An improvement in the performance of the pose estimators was observed by 47.6\% and 16.1\% upon integration of the motion blur and in-the-wild video modules respectively. Thus, the paper presents the following contributions.

\begin{enumerate}
    \item \emph{\textbf{A focused augmentation strategy that incorporates motion blur artifacts}}, which challenges the conventional belief in complex pipelines and shows significant improvements in handling these challenges.
    \item \emph{\textbf{Leveraging in-the-wild datasets}}, aids in capturing the variability and complexity present in the data, resulting in efficient and versatile pose estimation.
    % \item \emph{\textbf{A spatiotemporal approach to effectively synchronize partially synchronized frames}}, which employs a weighted cost function focusing on spatial and temporal joint positions reinforced by histogram representation, allowing the maintenance of temporal coherence.
    \item \emph{\textbf{Improved performance of existing pose estimators with proposed framework incorporation}}, where we demonstrate the substantial enhancement achieved by integrating our framework with existing pose estimation methods.
    % \item \emph{\textbf{A new dataset for player performance analysis for baseball sport}} which consists of more than 30,000 images and 20+ performance analysis metrics along with its corresponding 3D pose obtained from the Hawkeye camera system.
\end{enumerate}

The remainder of the paper is structured as follows. Section \ref{review}, provides a comprehensive review of previous research work on human pose modeling and sports analytics for vision. Section \ref{method} and \ref{sub:datasets} outline the methodology of the proposed work to tackle motion blur and the challenges present in the dataset respectively. The experimental setup and results are presented in Section \ref{exp}. Finally, Section \ref{conc} concludes the work done and addresses the potential improvements to the proposed approach.

\section{Related Works} \label{review}

\paragraph{\textbf{Vision for Sports Analytics}} In recent years, there have been significant advancements in sport-related pose estimation and body modeling techniques, which have greatly contributed to accurate performance analysis and understanding of human movement in sports. These techniques address common challenges faced in vision-based sports analytics, such as motion blur and occlusion. One approach, proposed by \cite{holistic-deblur}, presents a unified framework that combines deblurring and holistic 3D human body reconstruction. When the human reconstruction module is integrated, the deblurring module benefits from the human reconstruction loss, resulting in improved performance. Another approach, introduced by \cite{Lumentut_2020_ACCV}, focuses on a newly generated blurry human dataset and localized adversarial modules. Although these techniques have demonstrated significant improvements, they still encounter limitations, particularly in scenarios with dynamic backgrounds and significant motion differences between frames. Further research is needed to overcome these challenges and advance the effectiveness of vision-based sports analytics. 

%\textbf{Add a few more}.

\paragraph{\textbf{Human Pose Estimation}} Several methods have been proposed for 2D pose estimation, which can be broadly categorized into heatmap-based and regression-based approaches. Heatmap-based methods \cite{vitpose, mspn, hrnet} focus on predicting heatmaps that represent the likelihood of each keypoint being present at different locations in the image. These heatmaps are then processed to estimate the exact keypoint locations. On the other hand, regression-based methods \cite{panteleris2021peformer, poseur} directly regress the coordinates of keypoints from the input image by employing deep neural networks to learn the mapping between the image and the keypoints. These approaches have demonstrated impressive performance in capturing fine-grained details and handling occlusions, making them suitable for challenging pose estimation tasks. To estimate 3D poses from 2D keypoints, transformer-based networks have emerged as SOTA models. Epipolar Transformers \cite{epipolar} utilizes epipolar constraints to enforce geometric consistency between 2D keypoints. TransFuse \cite{transfuse} incorporates a cross-modal transformer to fuse information from multiple views. MHFormer \cite{mhformer} introduces multihead self-attention mechanisms to capture both local and global dependencies.

\begin{figure*}
{\centering
\includegraphics[width=17cm]{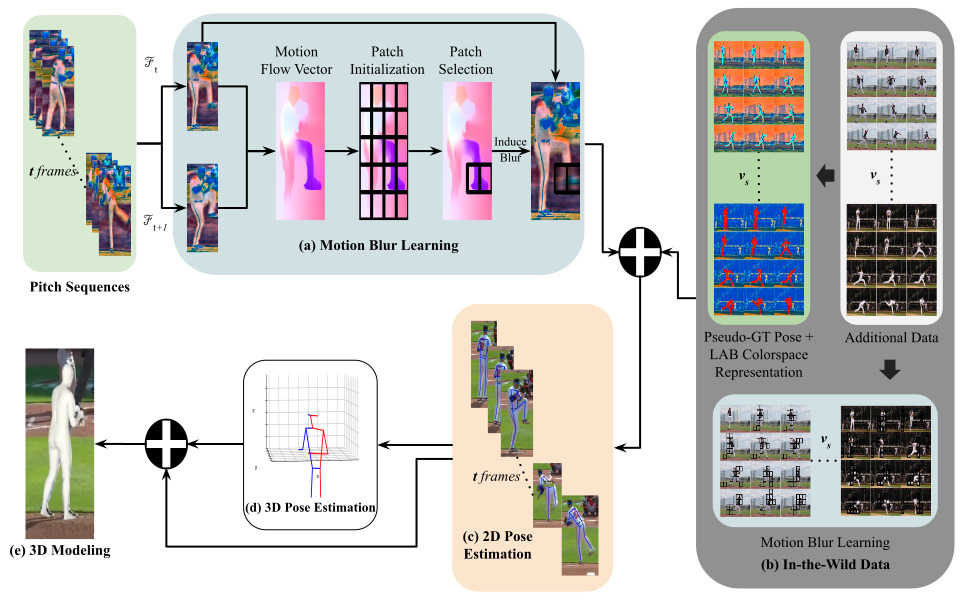}
\caption{Overview of the proposed system. (a) The motion blur learning module creates synthetic blur effects to the pitch sequences to learn better features and generalize well despite such effects. (b) In-the-wild data is leveraged to enhance the robustness of the model on diverse environmental conditions. (c) A regressor-based 2D pose estimator to train the data from the motion blur learning module and the in-the-wild data. (d) A transformer-based 3D pose estimator to train on sequential 2D pose data to estimate the 3D pose. (e) Concatenation of the 2D and 3D poses to estimate the 3D body mesh using spectral GCN.} 
\label{fig:arch}
}
\end{figure*}

\paragraph{\textbf{3D Pose and Shape Estimation}} Estimating both the 3D pose and shape of a human body from a single image is a challenging task with numerous applications in sports performance analysis. The SOTA methods in this area can be categorized into parameter-based and parameter-free approaches. Parameter-based methods, such as HMR \cite{HMR} and VIBE \cite{vibe}, utilize deep learning techniques to regress the parameters of a parametric body model from the input image. These methods typically employ convolutional neural networks (CNNs) to extract image features and then predict the model parameters from the features. However, the use of pose parameters as a regression target can introduce inaccuracies and non-minimal representations, leading to performance limitations \cite{noposep}. Thus, recent works use a parameter-free approach \cite{gcmr, pose2mesh}, aiming to directly regress the 3D coordinates of the mesh vertices without relying on a predefined parametric model. These methods typically often combine CNNs and graph convolutional networks (GCNs) to directly predict the 3D coordinates of the body mesh vertices. 

The parameter-free approach leverages the relationships and dependencies between body parts through graph structures, capturing contextual information to improve the accuracy and robustness of the human body model. Thus, in this work, we use a parameter-free approach inspired by Pose2Mesh \cite{pose2mesh} to directly regress the mesh coordinates from the 2D and 3D joint keypoints of the pitcher. 

% Parameter-based methods offer controllable and realistic representations of the human body, making them suitable for applications that require precise and customizable models. Parameter-free methods offer more flexibility and adaptability as they do not rely on predefined body models, allowing them to capture finer details and handle more diverse body shapes and poses.

% In summary, the advancements in 2D pose estimation, 3D pose estimation, and 3D pose and shape estimation have significantly contributed to the field of sport-related pose analysis. These techniques provide valuable insights into human movement, facilitating performance evaluation and enhancing training strategies in various domains including sports.

% \begin{figure*}
% {\centering
% \includegraphics[width=15cm]{Figures/architecture.png}
% \caption{Overview of the proposed system. (a) The Motion Blur Learning (MBL) module is used to create synthetic blur effects to the pitch sequences to help the network learn better features and generalize well despite such effects. (c) A regressor-based 2D pose estimator to train the data tweaked using the dataset augmented with the data from the MBL module (d) A transformer-based 3D pose estimator to train on sequential 2D pose data to estimate the 3D pose. (e) The 2D and 3D poses are concatenated and fed to a spectral graph convolution network to estimate the 3D body mesh of the pitcher.} 
% \label{fig:arch}
% }
% \end{figure*}

\section{Methodology} \label{method}

The proposed approach comprises several key steps aimed at enhancing the motion blur effect and estimating the 3D body model of the pitcher. Each pitch sequence is represented by a set $\hat{\mathcal{P}} = \{\mathcal{F}_t : \mathcal{F}_t \in \mathbb{R}^{H \times W \times 3}\}_{t=1}^{t_n}$. To augment the motion blur effect, the approach utilizes a motion blur learning module, where pairs of subsequent frames are taken for motion flow analysis. Each frame is divided into $k$ patches of equal size from which $\mathcal{N}$ patches are selected to induce motion blur. The motion flow vector $\mathcal{M}_k^{(t)}$ of each patch is then estimated, which is denoted as $\mathcal{M}_k^{{t}} = \sum_{i,j = 0, 0}^{i,j = H, W} v_{ij}$ where $v_{ij}$ is flow vector for each patch at pixel position $(i,j)$. Then, $\mathcal{N}$ patches with most $\mathcal{M}_k^{{t}}$ value is selected as the target regions to introduce motion blur effect. 

% The patch $\mathcal{I}_{\text{max}}$ with the highest concentration of motion flow vectors, represented as $\mathcal{I}_{\text{max}} = \max(\mathcal{I}_k)$, indicates regions of significant motion. Motion blur is then selectively induced to $\mathcal{I}_{max}$.

% In this module, pairs of subsequent frames, denoted as $\mathcal{F}_t$ and $\mathcal{F}_{t+1}$, are taken for motion flow analysis.

Next, the 2D pose of the pitcher in each frame $\mathcal{F}_t$ is estimated, where the input is a frame containing the pitcher, and the output is a pose representation denoted as $\mathcal{P}_{2D}^{(t)} \in \mathbb{R}^{\mathcal{J} \times 2}$, where $\mathcal{J}$ represents the total joints of the pitcher. Following the 2D pose estimation, the 3D pose of the pitcher is estimated by utilizing $s$ consecutive sets of 2D pose data as the receptive field, where the input is denoted as ${\mathcal{P}_{2D}} \in \mathbb{R}^{s \times \mathcal{J} \times 2} $. The output for the 3D pose estimation is a pose representation which is denoted as $\mathcal{P}_{3D} \in \mathbb{R}^{1 \times \mathcal{J} \times 3}$.

The 2D and 3D poses are then concatenated to form the input for the 3D body model, denoted as $\mathcal{P}_{\text{concat}}^{(t)} \in \mathbb{R}^{1 \times \mathcal{J} \times 5}$, which is the concatenation of its corresponding $\mathcal{P}_{2D}$ and $\mathcal{P}_{3D}$. The output of the 3D body model is represented as $\mathcal{{H}}_{3D} \in \mathbb{R}^{\mathcal{V} \times 3}$, where $\mathcal{V}$ represents human vertices of the mesh.

\subsection{Motion blur learning module} \label{mbl}

The motion blur learning module aims to address the motion blur challenges of the dataset by augmenting the dataset with extra synthetic data mimicking it in a realistic manner. It essentially provides the network with different opportunities to see and learn from different instances of motion blur, by increasing the frequency and consistency of the effects, thereby increasing the robustness of the challenge. 

To achieve realistic synthetic effects, our approach comprises a series of deliberate steps. Initially, we estimate the motion flow vectors between consecutive pairs of images. Subsequently, we integrate a two-step process to discern the specific regions (patches) where motion blur should be induced. This process involves a selective identification of patches that exhibit significant motion, ensuring a targeted approach to motion blur application.

The processed consecutive images from the dataset processing module (Section \ref{sec:processing-module}) are passed through a motion flow estimation algorithm proposed by \cite{gma}. It uses a transformer network to compute the attention matrix based on self-similarities to study the long-range dependencies between pixels of the same reference frame, which is then used to aggregate the motion features represented as shown in Equation \eqref{eq:gma} which is then augmented to the method proposed by RAFT \cite{raft}. This motion flow algorithm was utilized specifically, considering the fact that it can handle occluded regions well since it also considers the self-similarities between frames.

% An approach proposed by \cite{gma} was employed to study the motion flow between consecutive images. 

\begin{equation}
    \hat{\mathcal{F}}_t = \mathcal{F}_t + \lambda \sum_{j=1}^N A(\theta(\mathcal{x}_t), \phi(\mathcal{x}_j), \sigma(y_j)) \label{eq:gma}
\end{equation}

Here, $f_i$ represents motion features, $\theta$, $\phi$ and $\sigma$ denotes the projection of the query, key, and value, $\lambda$ denotes the learned parameter and $A$ denotes the self-similarity attention function.

The procedure for selectively inducing motion blur in specific regions involves a sequential two-step approach following the acquisition of motion flow vectors between consecutive frame pairs. This method comprises Patch Initialization and Patch Selection stages. In the Patch Initialization stage, individual images are partitioned into $4 \times 5$ patches. Subsequently, in the Patch Selection stage, a total of $\mathcal{N}$ patches are identified by ranking them based on the magnitude of their associated motion flow vectors. These selected patches are chosen as the targeted regions for inducing motion blur. Notably, this two-step process offers a distinct advantage by ensuring the introduction of motion blur is focused on regions that exhibit significant motion.

The process of applying motion blur to the chosen patches involves the utilization of a motion blur filter. This filter is inherently oriented and is parameterized by a rotation matrix. The angle of rotation ($\omega$), and the scale factor $s_f$ are key determinants of the filter's behavior. This oriented filter is centered at coordinates $(k_{s}/2, k_{s}/2)$, where $k_{s}$ denotes the kernel size. Notably, this filtering approach is applied to the selected $\mathcal{N}$ patches. The filtering procedure can be formally represented as:

\begin{equation}
\mathcal{B}(x, y) = \dfrac{\mathcal{R}}{\sum\mathcal{R}} \ast I_{\text{k}}(x, y) \label{eq:filter}
\end{equation}

where,
% \[
%     \mathcal{R} = \begin{bmatrix}
%     \cos(\omega) & -\sin(\omega) \\
%     \sin(\omega) & \cos(\omega)
%     \end{bmatrix}
% \]

\begin{equation}
    \mathcal{R} = \left[ \begin{smallmatrix}
    \cos(\omega) \cdot s_f & -\sin(\omega) \cdot s_f & (k_s//2) \cdot (1 - \cos(\omega) \cdot s_f) + (k_s//2) \cdot \sin(\omega) \cdot s_f \\
    \sin(\omega) \cdot s_f & \cos(\omega) \cdot s_f & (k_s//2) \cdot (1 - \cos(\omega) \cdot s_f) - (k_s//2) \cdot \cos(\omega) \cdot s_f
    \end{smallmatrix} \right]
\end{equation}

Here, $\mathcal{B}(x,y)$ represents the filtered pixel value at position (x,y) of the blurred patch. $I_{k}$ denotes the $k_{th}$ patch of the image and $(*)$ denotes the convolution operation between rotation matrix $\mathcal{R}$. The summation of $\mathcal{R}$ in the denominator of the fraction ensures normalization, preserving the intensity of the patch. This normalization prevents unwanted intensity changes and contributes to a realistic motion blur effect.

\subsection{In-the-Wild Video Integration} \label{sec:itw}

% Furthermore, to enhance the generalizability of the model, we took advantage of videos from various sources which capture the pitching action using slow-motion cameras with high resolution. Firstly, the pose of the pitcher was estimated in these videos which act as the pseudo-groundtruth for the corresponding frames. Then, the previously proposed blurring strategy in Subsection \ref{mbl} was employed to induce motion blur to fast-moving regions of the image to emulate the effect of the movements in low-quality videos. 

% The rationale behind this approach was to capitalize on publicly available videos on the internet to enhance the overall performance of our pose estimation system. Since our pose estimation method demonstrated proficiency in handling high-resolution and slow-motion videos, we used these videos to train our pose estimator where the pose obtained before blurring was used as GT pose data.

Furthermore, to enhance the generalizability and robustness of the pose estimator, we leveraged videos from various public sources, featuring slow-motion recordings of pitching actions in professional baseball games or practice sessions, captured at high resolution. This strategic inclusion of external data offered two pivotal advantages: diversity and an abundance of training data, both of which aided in capturing a wide spectrum of pitching scenarios, encompassing different players, camera angles, lighting conditions, and pitching styles. 

% This diversity was essential for the model to learn from a comprehensive array of real-world challenges and variations, enabling it to adapt effectively to a broad range of dynamic pitching environments.

Firstly, a diverse set of videos ($v_s$) were captured from publically available sources on the internet. These videos consisted of slow-motion videos of pitching action in professional baseball games or practice seasons captured at high resolution. Then, to train our pose estimator using these videos, we first estimated the pose of the pitcher in each frame, effectively generating pseudo-ground truth data for the corresponding frames. Next, to emulate the effects of movements and challenges often observed in low-quality videos, we employed the blurring strategy proposed in Section \ref{mbl}. This technique induced motion blur in all fast-moving regions of every image within the videos. By subjecting the model to this motion-blurred data, it learned to handle scenarios characterized by motion artifacts and low-quality video conditions, thus enhancing its resilience in practical settings.

Consequently, the motion blur-induced images, along with their corresponding pseudo-groundtruth data, were included in the training dataset alongside the existing data. This comprehensive training approach capitalized on the combination of diverse video sources, accurate pose estimation from high-quality videos, and exposure to motion blur-induced images, resulting in a pose estimator that demonstrated robustness and proficiency in estimating poses, especially under challenging conditions.

% By following this approach, we capitalized on publicly available videos from the internet to enhance the overall performance of our pose estimation system. The combination of diverse video sources, accurate pose estimation from high-quality videos, and the induction of motion blur allowed our model to become robust and effective in estimating poses even in challenging conditions, thus advancing the pose estimation in dynamic scenarios like pitching actions.

\subsection{Human Pose Estimation} \label{sec:pose-est}

Estimating the 3D body model of the pitcher offers several distinct advantages over traditional 2D and 3D pose estimation approaches. It facilitates comprehensive analyses, including the assessment of the pitcher's interaction with the environment, accurate computation of pitch trajectories and release points, and detailed biomechanical evaluation. By harnessing the power of 3D body model estimation, our understanding of the pitcher's movement patterns and biomechanics becomes significantly enriched, leading to valuable insights for optimizing performance and injury prevention strategies.

Thus, to estimate the 3D body model of pitchers in each frame, we first enhance the training data by incorporating synthetic artifacts proposed in Sections \ref{mbl} and \ref{sec:itw}. This augmented dataset is then used to train a regressor-based 2D pose estimator as described in the work by PEFormer \cite{panteleris2021peformer}. In contrast to SOTA estimators that rely on heatmaps, we opted for a regressor-based approach due to potential challenges in achieving accurate overlap between the 2D pose of the pitcher obtained through the optimization process of the camera projection (explained in Section \ref{camera-projection}). Subsequently, a vision transformer network proposed by MHFormer\cite{mhformer} is used to lift sequences of estimated 2D poses to 3D, resulting in the 3D pose for each corresponding input frame. The loss functions leveraged for 2D and 3D pose estimator is defined as:

\begin{equation}
    \mathcal{L}_{pose} = \frac{1}{\mathcal{N}} \sum_{i=1}^\mathcal{N} \frac{1}{\mathcal{J}} \sum_{j=1}^{\mathcal{J}} \| kp^{(ij)}_{pred} - kp^{(ij)}_{gt} \|_{\gamma}, 
    \label{eq:loss}
\end{equation}

where,
\[
\| \cdot \|_{\gamma} = 
\begin{cases}
    \| \cdot \|_2, & \text{if } \gamma = 2 \text{ (for $\mathcal{P}_{2D}$)} \\
    \| \cdot \|_3, & \text{if } \gamma = 3 \text{ (for $\mathcal{P}_{3D}$)}
\end{cases}
\]

and \(kp_{pred}^{(ij)}\) and \(kp_{gt}^{(ij)}\) corresponds to the estimated and ground truth pose from the pose estimator. $\| \cdot \|_{\gamma}$ denotes the Euclidean distance between the $
\gamma$ dimensional pose keypoints.  

$\mathcal{P}_{2D}$ and $\mathcal{P}_{3D}$ obtained for each frame are then concatenated into $\mathcal{P}_{concat} \in \mathbb{R}^{\mathcal{J} \times 5}$ and fed into a spectral convolution network \cite{specgcn} inspired by the works of Pose2Mesh \cite{pose2mesh}. The goal is to directly map the concatenated 2D and 3D poses to the body mesh of the pitcher. The loss function employed to train the mesh network is defined as:

\begin{equation}
\mathcal{L}_{mesh} = \lambda_v \mathcal{L}_{v} + \lambda_j \mathcal{L}_{j} + \lambda_n \mathcal{L}_n + \lambda_e \mathcal{L}_e
\label{loss-mesh}
\end{equation}

Here, $\mathcal{L}_v$ represents the $L_{1}$ distance between the output mesh and the GT mesh. $\mathcal{L}_j$ measures the loss between the 3D pose of the predicted mesh and the GT mesh. $\mathcal{L}_n$ corresponds to the loss of smoothness consistency and $\mathcal{L}_e$ denotes the loss of edge length consistency. The weights for each loss function are denoted by $\lambda$.

\section{Dataset} \label{sub:datasets}

% The availability of public datasets for 3D pose estimation in baseball sports analytics is limited. 
In this research, we use a comprehensive baseball dataset comprised of over 1000 games, including more than 100,000 pitches. The dataset encompasses 18 3D joint position information of pitchers, along with detailed pitch statistics such as pitching velocity, extension, pitch type, horizontal/vertical break, spin rate, pitch set, release velocity, and a few other observed pitch metrics. Additionally, it provides the 3D joint positions of both the batter and the catcher as well, facilitating a comprehensive analysis of the game dynamics.
% Fig. \ref{fig:dataset} shows some visualization of the cropped pitcher and its corresponding 3D poses captured from the Hawkeye system. 

In this work, to train the 2D and 3D pose estimators, we have used 150 pitch sequences, which correspond to 30,000 frames in total. Only 150 pitch sequences were used to strike a balance between having enough data to train a reliable model and avoiding the risk of overfitting by focusing on a smaller but more diverse and representative dataset. The dataset has been divided into three subsets: training, validation, and testing with the respective split provided in Table \ref{tab:datasplit}. 

% This enabled a comprehensive analysis of the player's posture. 

\begin{table}[H]
  \caption{Dataset split for Training, Validation, and Testing}
  \label{tab:datasplit}
  \centering
  \begin{tabular}{ccc}
    \toprule
    Dataset & Pitch Sequences & Frames\\
    \midrule
    Train & 105 & 21,050 \\
    Validation &  15 & 2,962\\
    Test & 30 & 5,988\\
    \bottomrule
  \end{tabular}
\end{table}

The size and composition of the dataset were carefully chosen after considering the diversity and coverage of pitching actions, including the handiness of the pitcher, the pitching set, the pitch type, and the lighting conditions of the game. Experimentation was conducted exclusively on the test set containing real-world input frames with inherent motion blur effects. We deliberately refrained from introducing synthetic motion blur to the test set to maintain the fidelity of the evaluation process by assessing under conditions that mirror real-world scenarios.

\begin{figure}[H]
{\centering
\includegraphics[width=8cm]{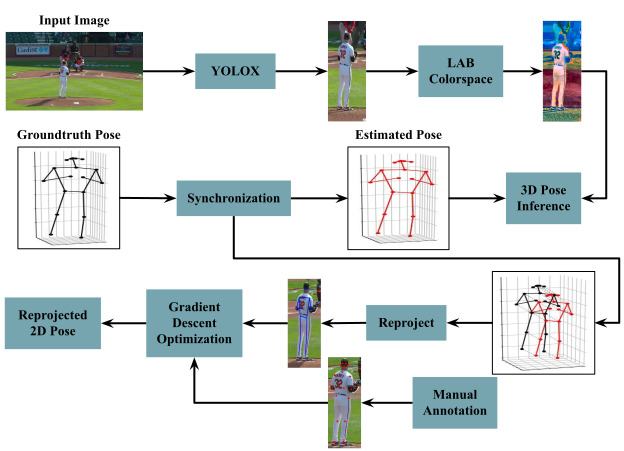}
\caption{Overview of the data processing to work around the limitations of the dataset.} 
\label{fig:dataset-module}
}
\end{figure}

Some problems with the dataset include extra frames before/ after the pitching action for which the pitcher's pose is not annotated and the absence of the camera parameters. Camera parameters are required to reproject and estimate the groundtruth 2D pose of the pitcher in camera coordinates from the annotated groundtruth 3D pose data captured in the world coordinate frame. This introduces some additional constraints to estimate the exact mapping parameters between the 3D world coordinates and 2D camera coordinates which is required to train the 2D pose estimator. Consequently, certain assumptions and approximations were made to work around the limitations in this dataset to ensure its validity, and those components can be visualized in Figure \ref{fig:dataset-module}. The proposed data processing module is extensively explained in the following subsection.

\subsection{Data processing module} \label{sec:processing-module}

The data processing module serves three primary objectives. Firstly, we tackle the task of enhanced player identification, focusing specifically on the pitcher. Leveraging advanced player detection techniques, allows us to accurately detect and isolate the pitcher in each frame while improving their visibility using enhancement techniques. Secondly, to ensure data consistency, frames with missing annotations are systematically removed, creating a clean and reliable dataset for further analysis. Lastly, we tackle the estimation of camera parameters, a critical step in obtaining groundtruth 2D pose information. By accomplishing these objectives, our data processing module becomes instrumental in preparing a high-quality and reliable dataset.
% , paving the way for robust and accurate pose estimation results.

\subsubsection{Pitcher Detection}

In this work, pitchers were the player of interest, so a pitcher detector network has been trained using YoloX \cite{yolox} to isolate the pitcher in each frame. The cropped frames with pitcher were then converted into the LAB colorspace which consists of two color channels and a luminosity channel. Then, contrast enhancement on the luminosity channel was done to increase the overall brightness level and amplify the adjacent pixel intensities. It was then converted to BGR colorspace to improve the visual quality of the images while still preserving color information.  

\subsubsection{Synchronization}

To synchronize between frames and the 3D GT pose data, Dynamic Time Warping (DTW) \cite{dtw} was employed as a method of aligning the two sequences. By warping the time axis and minimizing the distance or cost between the sequences, DTW finds the ideal alignment. Directly aligning an image with 3D keypoints is impractical; thus, a two-step approach was adopted. Firstly, an off-the-shelf 3D pose estimator was used to estimate the pose at each frame. Subsequently, DTW was used to find the best alignment between the GT 3D pose sequence and the estimated poses. Given the data and the constraints of the problem, a one-to-one relation was established as a hard constraint thereby enforcing unique correspondences between poses. Equation \eqref{eq:dtw} in the following represents the cost function $(\mathcal{G})$ that was constructed as part of the alignment process and takes into account both the spatial and temporal components of the data. 

\begin{equation}
\begin{aligned}
\mathcal{G} = & g_s \left(\frac{1}{\mathcal{J}} \sum_{i=1}^{\mathcal{J}} (kp_{gt}^{(i)} - kp_{pred}^{(i)})^2\right) \\
& + g_t \left(1 - \frac{\sum_{i=1}^{\mathcal{J}} kp_{gt}^{(i)} \cdot kp_{pred}^{(i)}}{\sqrt{\sum_{i=1}^{\mathcal{J}} (kp_{gt}^{(i)})^2} \cdot \sqrt{\sum_{i=1}^{\mathcal{J}} (kp^{(i)}_{pred})^2}}\right)
\end{aligned}
\label{eq:dtw}
\end{equation}

% \begin{equation}
%     \mathcal{G} = hist(g_s s_{cost} + g_t  t_{cost}) \label{eq:dtw}
% \end{equation}

% here,
% \begin{equation}
%     s_{cost}  = mse(kp_{hw}, kp_{est}) \label{mse}
% \end{equation}
% \begin{equation}
%     t_{cost} = cosine\_sim(kp_{hw}, kp_{est}) \label{cosine}
% \end{equation}

Here, $kp_{gt}$ and $kp_{pred}$ are the GT keypoints and estimated keypoints respectively, and $g_s$ and $g_t$ correspond to the spatial and temporal weight gains, respectively. $\mathcal{G}$ is formulated to simultaneously account for spatial and temporal aspects of the pose data, providing a more accurate approach to align the data. Mean square error was used to capture the spatial discrepancy between keypoints and cosine similarity was utilized to measure the difference in angle between subsequent frames to calculate the temporal context of the pose data. The estimated spatial and temporal distances are then added and represented as bins of a histogram to then compare with other pose representations. 

\subsubsection{Camera Projection} \label{camera-projection}

To address the absence of camera parameters for mapping 3D world pose coordinates to 2D camera pose coordinates, we utilize an iterative optimization approach to find the optimized camera parameters. We begin by manually annotating the 2D pose in a reference frame and initializing a focal length. Through a process of gradient descent optimization, we iteratively refine the mapping by adjusting the focal length. This adjustment is performed to minimize the error between the projected 2D pose and the annotated 2D pose as shown in Equation \eqref{eq:loss}. Equation \eqref{eq:gd} outlines the optimization process used in this approach. 

\begin{equation}
    \hat{f} = f_{i} - \alpha \Delta L(f_{i}) \label{eq:gd}
\end{equation}

Here, $\hat{f}$ and $f_{i}$ represent the updated and previous focal lengths respectively, and $\alpha$ and $\Delta L(f_{i})$ represent the learning rate and the gradient of the loss function respectively. 

During each iteration, the gradient of the loss function with respect to the focal length is computed. This gradient guides the adjustment of the focal length towards values that result in a more accurate 2D pose projection. By iteratively updating the focal length in the direction that reduces the loss, the mapping between 3D world coordinates and 2D camera coordinates is progressively refined. This iterative optimization scheme enhances the accuracy of the 3D-to-2D projection by effectively incorporating camera parameters.
% Although this method relies on assumptions and iterative optimization, it provided to be a good baseline for estimating the 2D poses without any camera parameter information.

\section{Experimentation} \label{exp}

\subsection{Implementation Details}

The training process was conducted on a system equipped with an Intel i7 processor (16 cores, 16 GB RAM) and an Nvidia 3050Ti GPU with 4GB of dedicated RAM. For the training process, the dataset was split according to the specifications mentioned in the subsection \ref{sub:datasets}. Both pose estimator models underwent training for 100 epochs, utilizing a batch size of 16 along with a total of 16 workers for concurrent processing.

For 2D pose estimation, a cross-covariance encoder \cite{xcit} was employed, along with a simple transformer decoder \cite{detr}. The input image was divided into patches of size $16\times16$, which were then flattened into tokens. The AdamW optimizer was used with a weight decay of $10^{-4}$, and the learning rate was set to $10^{-5}$ for the encoder and $10^{-4}$ for the decoder. For 3D pose estimation, the Adam optimizer was utilized along with a Reduce Learning Rate on the Plateau scheduler. The scheduler had a patience of 5 and reduced the learning rate by a factor of 0.3. The sequence length was fixed at 27. The mean per joint position error (MPJPE) \cite{mpjpe} was employed as a loss function for the pose evaluation as shown in Equation \eqref{eq:loss}. GT mesh models were not available for training the 3D models.  Thus, pseudo-groundtruths were generated using the method described in \cite{smplifyx}. The learning rate (LR) was initialized as $10^{-4}$, and a multistep LR scheduler was used with a LR factor of 0.1. The RMSProp \cite{rmsprop} optimizer was used to optimize the model during training. 
% Additionally, a multitask loss function was utilized, as shown in Eq. \eqref{loss-mesh}. 

\subsection{Motion Blur Learning}

\begin{figure*}
{\centering
\includegraphics[width=16.5cm]{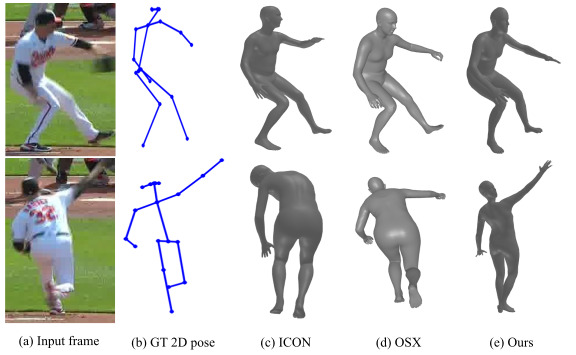}
\caption{Qualitative evaluation of pose reconstruction and the effectiveness of the proposed framework in handling motion blur effects.}  
\label{fig:smpl}
}
\end{figure*}

A thorough comparative analysis against existing approaches can be visualized from Figure \ref{fig:smpl}, which highlights the substantial advancements achieved through our proposed method, particularly evident during pitching actions in the second row. The results demonstrate a notable enhancement in the 3D body model, reaffirming the effectiveness and superiority of our approach.

To strike a balance between augmentation and the complexity of the augmenting task, it is pivotal to avoid over-augmentation, as it can lead to overfitting of the network and hinder performance on unseen data. To determine the optimal hyperparameters, we performed three additional experiments, the results of which are presented in Tables \ref{tab:kernels} - \ref{tab:patch-type}. 

% Figure \ref{fig:smpl} demonstrates the qualitative visualization of the performance of our module when compared to some existing works. Significant improvement iSignificant improvement in the 3D body model upon using the proposed approach has been observed especially while pitching action (in the second row). n the 3D body model upon using the proposed approach has been observed especially while pitching action (in the second row). 

\subsubsection{Varying the number of filters}

We aimed to find the optimal number of filters that could accurately simulate realistic motion blur effects. By varying the number of filters used, we assessed the performance of the pose estimator as shown in Table \ref{tab:kernels}.

% and identified the point at which the induced motion blur effect closely resembled real-world scenarios. This analysis helped us determine the most suitable number of kernels to be used for generating motion-blurred images during augmentation.

\begin{table}[H]
  \caption{Ablation study on varying number of filters for motion blur effect.}
  \label{tab:kernels}
  \centering
  \begin{tabular}{cc}
    \toprule
    Filters & Loss\\
    \midrule
    0 & 1.15\\
    1 & 0.68\\
    \pmb{2} & \pmb{0.55} \\
    3 & 1.43 \\
    4 & 2.28 \\
    5 & 3.44 \\
  \bottomrule
  \end{tabular}
\end{table}

% Table \ref{tab:kernels} illustrates the relationship between the number of kernels and the corresponding test loss of the 2D pose estimator, indicating the influence of motion blur in the pitch sequences. 

The increasing test loss beyond a certain point in Table \ref{tab:kernels} indicates that the network encounters difficulties in extracting informative features necessary for accurate pose estimation as the degree of motion blur intensifies. This shows that an excessive number of blur filters will lead to a diminishing capacity of the network to effectively handle and interpret motion blur in images.
% , resulting in a loss of pose estimation accuracy.

\subsubsection{Different Patch Size}

We conducted a study to investigate the impact of different patch sizes on the performance of the pose estimator. By varying the patch sizes ($s_{patch}$) and the number of patches ($\mathcal{N}$) in the input frames, we assessed how these factors influenced the accuracy of the pose estimator. This experiment enabled us to discover the best $s_{patch}$ and $\mathcal{N}$ for optimal performance. 
% Here, $\mathcal{N}$ doesn't imply random $n$ patches, but rather the fastest moving $n$ patches as discussed in Section \ref{mbl}.   

\begin{table}[H]
  \caption{Ablation study on the region size and frequency of motion blur effect}
  \label{tab:patch-size}
  \centering
  \begin{tabular}{cccccc}
    \toprule
    \diagbox{$s_{patch}$}{$\mathcal{N}$} & 1 & 3 & 5 & 7 & 9\\
    \midrule
    10 & 0.83 & 0.74 & 0.66 & 0.64 & 0.67 \\
    20 & 0.71 & 0.57 & 0.62 & 0.60 & 0.62 \\
    30 & 0.68 & \pmb{0.55} & 0.61 & 0.639 & 0.59 \\
    40 & 0.74 & 0.63 & 0.68 & 0.75 & 0.78 \\
    50 & 0.77 & 0.75 & 0.71 & 0.83 & 0.97 \\
    \bottomrule
  \end{tabular}
\end{table}

The extensive study conducted in \ref{tab:patch-size} demonstrates that the optimal results were achieved when using three patches, each with a patch size of 30. This finding aligns with qualitative observations, as this particular hyperparameter setup resulted in the most realistic blur pattern. 
% The use of multiple patches with an appropriate patch size allowed for capturing a more comprehensive representation of the motion blur present in the pitch sequences.

% Consequently, this setup leads to improved performance in estimating the pose and enhances the overall realism of the reconstructed images. The alignment between the quantitative results and qualitative assessment further reinforces the validity and effectiveness of the chosen hyperparameters for achieving high-quality and visually appealing results.

\subsubsection{Different Patch Types}

To identify the most suitable patch type that closely resembles a realistic representation of the motion blur effect, we conducted experiments involving different patch types. The results of this experiment are summarized in Table \ref{tab:patch-type}. 

% By evaluating various patch types, we aimed to assess their ability to accurately simulate the visual characteristics associated with motion blur. Each patch type was analyzed in terms of its impact on the overall performance of the pose estimation system.

\begin{table}[H]
  \caption{Comparison with different patch types}
  \label{tab:patch-type}
  \centering
  \begin{tabular}{cc}
    \toprule
    Patch Type & Loss\\
    \midrule
    None & 1.15 \\
    Binary Mask & 2.12  \\
    Inpainting & 1.57  \\
    Gaussian Blur & 0.99  \\
    \pmb{Motion Blur} & \pmb{0.55} \\
    \bottomrule
  \end{tabular}
\end{table}

% In our research experimentation, we conducted a comprehensive comparison of different masking techniques to determine their effectiveness in replicating the realistic motion of fast-moving objects. Among the various techniques evaluated, 

As shown in the results in Table \ref{tab:patch-type}, motion blur filters emerged as the superior method by successfully capturing the essence of rapid motion in a more accurate manner. 

Taking into account the findings from Table \ref{tab:kernels} to \ref{tab:patch-type}, the most optimal setup was with the adoption of a motion blur patch type that utilizes 2 filters, a patch size of 30, and 3 patches. This combination of parameters has demonstrated a greater ability to provide a substantial representation of the data, thereby significantly enhancing the generalizability of the pose estimator in handling motion blur effects. This finding contributes to effective pose estimation by handling challenging scenarios characterized by fast-moving actions in the images.

% By employing this optimized configuration, the pose estimator is expected to achieve improved performance in accurately estimating poses even in the presence of motion blur.

\subsection{Human Pose Estimation}

\subsubsection{Different data modules}

To evaluate the effectiveness of our method, we conducted tests on the curated dataset as described in Section \ref{sub:datasets}. Specifically, we evaluated the performance of our 2D and 3D pose estimation algorithms after augmenting the dataset with motion blur (MB) effects and in-the-wild (ItW) videos. The results of these evaluations are summarized in Table \ref{tab:modules}.

\begin{table}[H]
  \caption{Results of the estimated pose with different modules for training.}
  \label{tab:modules}
  \centering
  \begin{tabular}{ccccc}
    \toprule
    Base Model & ItW & MB & 2D Loss & 3D Loss\\
    \midrule
    \ding{51} &  &  & 1.05 & 1.93\\
    \ding{51} & \ding{51} &  & 0.88 & 1.61\\
    \ding{51} &  & \ding{51} & 0.55 & 1.47\\
    \ding{51} & \ding{51} & \ding{51} & \pmb{0.48} & \pmb{1.23}  \\
    \bottomrule
  \end{tabular}
\end{table}

% The experimental results demonstrate the effectiveness of the proposed approach in handling motion blur and generalizing to diverse datasets. 

By incorporating both ItW data and MB modules, the performance of the base model is significantly improved. The 2D loss shows a substantial improvement of 58\%, indicating enhanced accuracy in estimating the 2D pose of the human body. Subsequently, this improves the performance of the 3D pose estimator by 36\%.
% These results emphasize the model's ability to handle the challenges posed by motion blur and its generalizability to different types of data, ultimately leading to improved performance in pose estimation tasks.

\subsubsection{Comparison on SOTA pose estimators}

The experimental evaluation in Table \ref{tab:sota2d} demonstrates the performance improvement achieved by incorporating our approach during the training of SOTA 2D pose estimators using our dataset. Our objective was to show the efficacy of our approach in improving the overall performance of the pose estimators.

\begin{table}[H]
  \caption{Performance of different SOTA 2D pose estimation approaches with the proposed MB learning module. }
  % $\dagger$- With the proposed MB module}
  \label{tab:sota2d}
  \centering
  \begin{tabular}{cccc}
    \toprule
    Method & Type & MB & Loss \\
    \midrule
    Xu et al \cite{vitpose} & Heatmap &  & 1.37\\
    Ke et al \cite{hrnet} & Heatmap &  & 1.46 \\
    Panteleris et al\cite{panteleris2021peformer} & Regressor &  & 1.15 \\
    Li et al. \cite{mspn} & Heatmap &  & 1.83\\
    Mao et al. \cite{poseur} & Regression &  & 1.26 \\
    
    \midrule
    
    Xu et al \cite{vitpose} & Heatmap & \ding{51} & 1.17 (+0.20)\\
    Ke et al \cite{hrnet} & Heatmap & \ding{51} & 1.21 (+0.25) \\
    Panteleris et al\cite{panteleris2021peformer} & Regressor & \ding{51} & 0.55 (+0.60)\\
    Li et al. \cite{mspn} & Heatmap & \ding{51} & 1.46 (+0.37)\\
    Mao et al. \cite{poseur} & Regressor & \ding{51} & 0.61 (+0.65) \\
    \bottomrule
  \end{tabular}
\end{table}

The results indicate a significant improvement in the pose estimators' performance after integrating with the proposed approach, primarily due to its ability to handle motion blur. Furthermore, the comparison between heatmap-based and regression-based techniques highlights the limitations of the former in addressing the challenges of our dataset discussed in Section \ref{sec:pose-est}.

\section{Conclusion} \label{conc}

The research proposes a unique approach to accurately estimate the pose of pitchers in baseball games by addressing the challenges posed by the motion blur effect. An innovative augmentation technique has been proposed to increase the frequency and consistency of motion blur in a strategic pattern to enhance the network's ability to learn and adapt to these effects. Integrating in-the-wild video data into the training module with psuedo-groundtruth pose information aided the network to be effective against the variable lighting and camera positions. By training a 2D and 3D pose estimator on these data, significant improvements in the accuracy of pose estimation have been observed, particularly during pitching actions. Thus, this approach demonstrates a more focused and strategic augmentation strategy to induce motion blur can yield improvements in pose estimation emphasizing the importance of thoughtful augmentation in addressing the motion blur effect and offering an alternative perspective to traditional complex pipelines.

Future research directions include improving the quality of the body model by incorporating shape information and other visual contexts like background cues and scene understanding. Another focus area is synthesizing multiview camera data to gain richer information, resulting in more accurate and robust pose estimation.

\section{Acknowledgment}
This work was supported by the Baltimore Orioles, Major League Baseball through the Mitacs Accelerate Program. We are grateful for their support and contributions, which have significantly enriched the research and its findings. We also acknowledge Digital Research
Alliance of Canada for hardware support.
% for their support and collaboration in providing the necessary data and resources for conducting this research. 

\bibliographystyle{ieeetr}
\bibliography{egbib}

\begin{thebibliography}{10}

\bibitem{baseball-review}
K.~Koseler and M.~Stephan, ``Machine learning applications in baseball: A
  systematic literature review,'' {\em Applied Artificial Intelligence},
  vol.~31, pp.~1--19, 02 2018.

\bibitem{baseball1}
G.~Sidle and H.~Tran, ``Using multi-class classification methods to predict
  baseball pitch types,'' {\em J. Sports Anal.}, vol.~4, pp.~85--93, Feb. 2018.

\bibitem{baseball2}
C.~T. Heaton and P.~Mitra, ``Learning to describe player form in the {MLB},''
  {\em CoRR}, vol.~abs/2109.05280, 2021.

\bibitem{youtube}
MLB, ``Red sox vs. orioles game highlights (4/8/21) | mlb highlights.'' YouTube
  video, 2021.

\bibitem{blurcam1}
S.~Cho, Y.~Matsushita, and S.~Lee, ``Removing non-uniform motion blur from
  images,'' in {\em 2007 IEEE 11th International Conference on Computer
  Vision}, pp.~1--8, IEEE, 2007.

\bibitem{blurcam2}
T.~Sieberth, R.~Wackrow, and J.~Chandler, ``Motion blur disturbs--the influence
  of motion-blurred images in photogrammetry,'' {\em The Photogrammetric
  Record}, vol.~29, no.~148, pp.~434--453, 2014.

\bibitem{blurcam3}
A.~Agrawal, Y.~Xu, and R.~Raskar, ``Invertible motion blur in video,'' in {\em
  ACM SIGGRAPH 2009 papers}, pp.~1--8, 2009.

\bibitem{zhao2023human}
Y.~Zhao, D.~Rozumnyi, J.~Song, O.~Hilliges, M.~Pollefeys, and M.~R. Oswald,
  ``Human from blur: Human pose tracking from blurry images,'' {\em arXiv
  preprint arXiv:2303.17209}, 2023.

\bibitem{Lumentut_2020_ACCV}
J.~S. Lumentut, J.~Santoso, and I.~K. Park, ``Human motion deblurring using
  localized body prior,'' in {\em Proceedings of the Asian Conference on
  Computer Vision}, 2020.

\bibitem{holistic-deblur}
J.~Santoso, I.~K. Park, {\em et~al.}, ``Holistic 3d body reconstruction from a
  blurred single image,'' {\em IEEE Access}, vol.~10, pp.~115399--115410, 2022.

\bibitem{vitpose}
Y.~Xu, J.~Zhang, Q.~Zhang, and D.~Tao, ``Vitpose: Simple vision transformer
  baselines for human pose estimation,'' {\em Advances in Neural Information
  Processing Systems}, vol.~35, pp.~38571--38584, 2022.

\bibitem{mspn}
W.~Li, Z.~Wang, B.~Yin, Q.~Peng, Y.~Du, T.~Xiao, G.~Yu, H.~Lu, Y.~Wei, and
  J.~Sun, ``Rethinking on multi-stage networks for human pose estimation,''
  {\em arXiv preprint arXiv:1901.00148}, 2019.

\bibitem{hrnet}
K.~Sun, B.~Xiao, D.~Liu, and J.~Wang, ``Deep high-resolution representation
  learning for human pose estimation,'' in {\em Proceedings of the IEEE/CVF
  conference on computer vision and pattern recognition}, pp.~5693--5703, 2019.

\bibitem{panteleris2021peformer}
P.~Panteleris and A.~Argyros, ``Pe-former: Pose estimation transformer,'' in
  {\em International Conference on Pattern Recognition and Artificial
  Intelligence}, pp.~3--14, Springer, 2022.

\bibitem{poseur}
W.~Mao, Y.~Ge, C.~Shen, Z.~Tian, X.~Wang, Z.~Wang, and A.~v. den Hengel,
  ``Poseur: Direct human pose regression with transformers,'' in {\em European
  Conference on Computer Vision}, pp.~72--88, Springer, 2022.

\bibitem{epipolar}
Y.~He, R.~Yan, K.~Fragkiadaki, and S.-I. Yu, ``Epipolar transformers,'' in {\em
  Proceedings of the ieee/cvf conference on computer vision and pattern
  recognition}, pp.~7779--7788, 2020.

\bibitem{transfuse}
H.~Ma, L.~Chen, D.~Kong, Z.~Wang, X.~Liu, H.~Tang, X.~Yan, Y.~Xie, S.-Y. Lin,
  and X.~Xie, ``Transfusion: Cross-view fusion with transformer for 3d human
  pose estimation,'' {\em arXiv preprint arXiv:2110.09554}, 2021.

\bibitem{mhformer}
W.~Li, H.~Liu, H.~Tang, P.~Wang, and L.~Van~Gool, ``Mhformer: Multi-hypothesis
  transformer for 3d human pose estimation,'' in {\em Proceedings of the
  IEEE/CVF Conference on Computer Vision and Pattern Recognition (CVPR)},
  pp.~13147--13156, 2022.

\bibitem{HMR}
A.~Kanazawa, M.~J. Black, D.~W. Jacobs, and J.~Malik, ``End-to-end recovery of
  human shape and pose,'' in {\em Proceedings of the IEEE conference on
  computer vision and pattern recognition}, pp.~7122--7131, 2018.

\bibitem{vibe}
M.~Kocabas, N.~Athanasiou, and M.~J. Black, ``Vibe: Video inference for human
  body pose and shape estimation,'' in {\em Proceedings of the IEEE/CVF
  conference on computer vision and pattern recognition}, pp.~5253--5263, 2020.

\bibitem{noposep}
N.~Kolotouros, G.~Pavlakos, and K.~Daniilidis, ``Convolutional mesh regression
  for single-image human shape reconstruction,'' {\em Proceedings of the
  IEEE/CVF Conference on Computer Vision and Pattern Recognition},
  pp.~4501--4510, 2019.

\bibitem{gcmr}
N.~Kolotouros, G.~Pavlakos, and K.~Daniilidis, ``Convolutional mesh regression
  for single-image human shape reconstruction,'' in {\em Proceedings of the
  IEEE/CVF Conference on Computer Vision and Pattern Recognition},
  pp.~4501--4510, 2019.

\bibitem{pose2mesh}
H.~Choi, G.~Moon, and K.~M. Lee, ``Pose2mesh: Graph convolutional network for
  3d human pose and mesh recovery from a 2d human pose,'' {\em Computer
  Vision--ECCV 2020: 16th European Conference, Glasgow, UK, August 23--28,
  2020, Proceedings, Part VII 16}, pp.~769--787, 2020.

\bibitem{gma}
S.~Jiang, D.~Campbell, Y.~Lu, H.~Li, and R.~Hartley, ``Learning to estimate
  hidden motions with global motion aggregation,'' in {\em Proceedings of the
  IEEE/CVF International Conference on Computer Vision}, pp.~9772--9781, 2021.

\bibitem{raft}
Z.~Teed and J.~Deng, ``Raft: Recurrent all-pairs field transforms for optical
  flow,'' in {\em Computer Vision--ECCV 2020: 16th European Conference,
  Glasgow, UK, August 23--28, 2020, Proceedings, Part II 16}, pp.~402--419,
  Springer, 2020.

\bibitem{specgcn}
J.~Bruna, W.~Zaremba, A.~Szlam, and Y.~LeCun, ``Spectral networks and locally
  connected networks on graphs,'' {\em arXiv preprint arXiv:1312.6203}, 2013.

\bibitem{yolox}
Z.~Ge, S.~Liu, F.~Wang, Z.~Li, and J.~Sun, ``Yolox: Exceeding yolo series in
  2021,'' {\em arXiv preprint arXiv:2107.08430}, 2021.

\bibitem{dtw}
H.~Sakoe and S.~Chiba, ``Dynamic programming algorithm optimization for spoken
  word recognition,'' {\em IEEE Transactions on Acoustics, Speech, and Signal
  Processing}, vol.~26, no.~1, pp.~43--49, 1978.

\bibitem{xcit}
A.~Ali, H.~Touvron, M.~Caron, P.~Bojanowski, M.~Douze, A.~Joulin, I.~Laptev,
  N.~Neverova, G.~Synnaeve, J.~Verbeek, {\em et~al.}, ``Xcit: Cross-covariance
  image transformers,'' {\em Advances in neural information processing
  systems}, vol.~34, pp.~20014--20027, 2021.

\bibitem{detr}
K.~Li, S.~Wang, X.~Zhang, Y.~Xu, W.~Xu, and Z.~Tu, ``Pose recognition with
  cascade transformers,'' in {\em Proceedings of the IEEE/CVF conference on
  computer vision and pattern recognition}, pp.~1944--1953, 2021.

\bibitem{mpjpe}
C.~Ionescu, D.~Papava, V.~Olaru, and C.~Sminchisescu, ``Human3.6m: Large scale
  datasets and predictive methods for 3d human sensing in natural
  environments,'' {\em IEEE Transactions on Pattern Analysis and Machine
  Intelligence}, vol.~36, no.~7, pp.~1325--1339, 2014.

\bibitem{smplifyx}
G.~Pavlakos, V.~Choutas, N.~Ghorbani, T.~Bolkart, A.~A. Osman, D.~Tzionas, and
  M.~J. Black, ``Expressive body capture: 3d hands, face, and body from a
  single image,'' in {\em Proceedings of the IEEE/CVF conference on computer
  vision and pattern recognition}, pp.~10975--10985, 2019.

\bibitem{rmsprop}
G.~E. Hinton, ``Neural networks for machine learning, lecture 6a: Overview of
  mini-batch gradient descent,'' {\em University of Toronto}, 2012.

\end{thebibliography}

\end{document}